\documentclass[10pt,twocolumn,letterpaper]{article}

\usepackage{cvpr}

\usepackage{graphicx}
\usepackage{amsmath}
\usepackage{amssymb}
\usepackage{booktabs}
\usepackage{mathrsfs}
\usepackage{inputenc}
\usepackage{multirow}
\usepackage{booktabs}
\usepackage{color}
\usepackage{tabularx}
\usepackage{algorithm} 
\usepackage{algorithmic} 

\usepackage{flushend}

\usepackage{colortbl}

\usepackage[pagebackref,breaklinks,colorlinks]{hyperref}

\usepackage[capitalize]{cleveref}
\crefname{section}{Sec.}{Secs.}
\Crefname{section}{Section}{Sections}
\Crefname{table}{Table}{Tables}
\crefname{table}{Tab.}{Tabs.}

\begin{document}
\bibliographystyle{unsrt}
\title{From Less to More: Spectral Splitting and Aggregation Network for Hyperspectral Face Super-Resolution}

\author{
    Junjun Jiang$\,^{\dagger}$, Chenyang Wang$\,^{\dagger}$, Xianming Liu$\,^{\dagger}$, Kui Jiang$\,^\ddag$, Jiayi Ma$\,^\ddag$\\
    $\,^{\dagger}$ Harbin Institute of Technology\quad
    $\,^\ddag$ Wuhan University\\
}


\maketitle

\begin{abstract}
High-resolution (HR) hyperspectral face image plays an important role in face related computer vision tasks under uncontrolled conditions, such as low-light environment and spoofing attacks. However, the dense spectral bands of hyperspectral face images come at the cost of limited amount of photons reached a narrow spectral window on average, which greatly reduces the spatial resolution of hyperspectral face images.
In this paper, we investigate how to adapt the deep learning techniques to hyperspectral face image super-resolution (HFSR), especially when the training samples are very limited. Benefiting from the amount of spectral bands, in which each band can be seen as an image, we present a spectral splitting and aggregation network (SSANet) for HFSR with limited training samples. In the shallow layers, we split the hyperspectral image into different spectral groups. Then, we gradually aggregate the neighbor bands at deeper layers to exploit spectral correlations. By this spectral splitting and aggregation strategy (SSAS), we can divide the original hyperspectral image into multiple samples (\emph{from less to more}) to support the efficient training of the network and effectively exploit the spectral correlations among spectrum. To cope with the challenge of small training sample size (S3) problem, we propose to expand the training samples by a self-representation model and symmetry-induced augmentation. Experiments show that SSANet can well model the joint correlations of spatial and spectral information. By expanding the training samples, SSANet can effectively alleviate the S3 problem. 
\end{abstract}

\section{Introduction}
Benefit from hyperspectral imagery which can capture the local spectral properties of human tissue, hyperspectral face analysis has attracted more and more attention from scholars in the field of face related computer vision tasks because of its satisfactory performance under uncontrolled conditions, such as low-light environment and spoofing attacks. However, the hyperspectral face imaging system is often compromised due to the limitations of the amount of the incident energy. There is always a tradeoff between the spatial and spectral resolution of the real imaging process. With the increase of the spectral bands, all other factors kept constant, to ensure a high signal-to-noise ratio (SNR) the spatial resolution will inevitably become a victim. Therefore, how to obtain a reliable hyperspectral face image with high spatial resolution remains a very challenging problem.

\begin{figure*}
  \centering
  \includegraphics[width=0.98\linewidth]{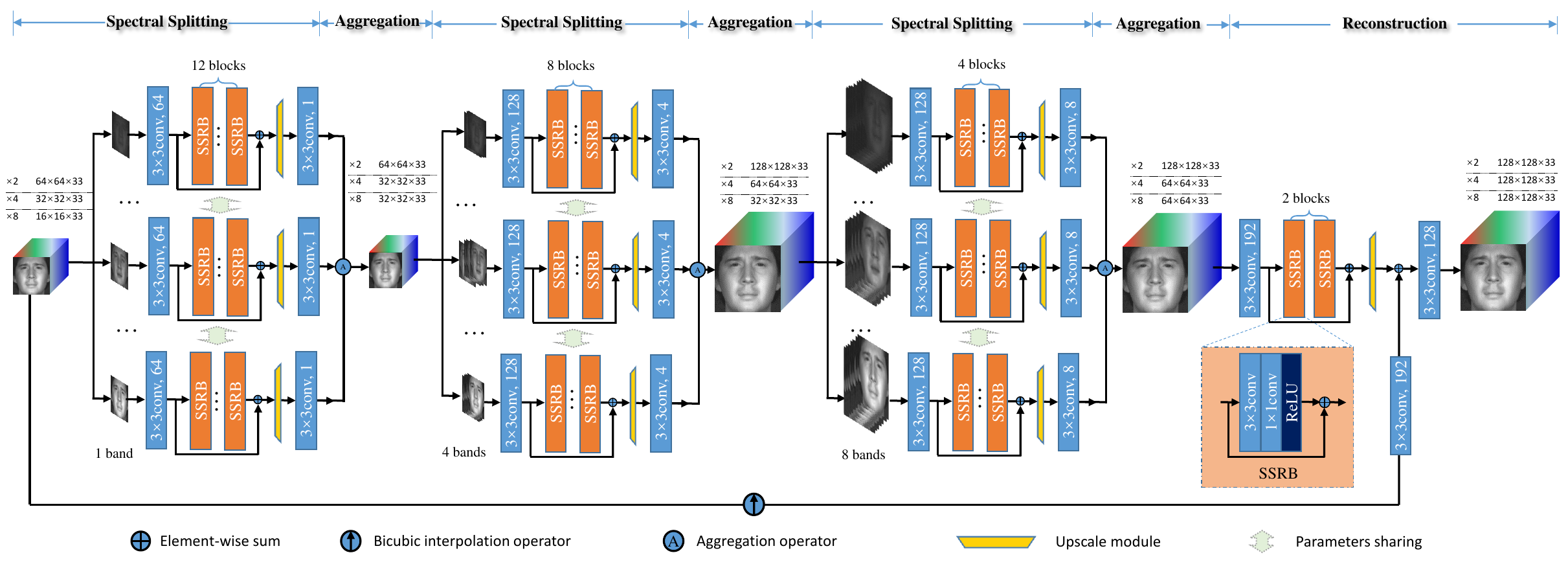}\\
  \caption{The flow chart of our proposed spectral splitting and aggregation network (SSANet) based hyperspectral face super-resolution (HFSR) method. SSANet includes three spectral splitting and aggregation modules followed by a reconstruction module. }\label{fig:Network}
\end{figure*}

Super-Resolution (SR) reconstruction can infer a high-resolution (HR) image from a single low-resolution (LR) image or sequential observed LR images \cite{Park2003}. It is a post-processing technique that does not require hardware modifications, and thus could break through the limitations of the imaging systems. Since the pioneer work of Bake and Kanda \cite{Baker2000}, face SR, \emph{a.k.i}, face hallucination, has received increased attention \cite{Liu2001, Wang2005Eig, Zhuang2007, Park2008}. Especially in recent years, the emergence of deep learning technology has promoted a large number of face SR methods \cite{jiang2023deep}. However, these methods are mainly for gray/RGB face images and cannot exploit the spectral correlation efficiently. This is mainly due to that the spectral dimensionality of hyperspectral face image is very high and the number of training samples of current hyperspectral face image dataset is extremely small, \emph{e.g.}, around 100.  There are not enough training samples to support the training of the complex deep network. In addition, these deep learning algorithms require large amounts of memory and are computationally expensive. 
Whether it is a traditional shallow learning-based method or a deep learning-based method, it is difficult to construct a powerful representation model from only dozens of images to reconstruct the target HR hyperspectral face images.

In summary, when we apply the existing learning-based face SR methods to hyperspectral images, the challenges are twofold: (i) the spectral correlations of hyperspectral face images cannot be fully utilized; (ii) hyperspectral face images are difficult to obtain, and the training samples of existing hyperspectral face image database is very small, thus cannot support these complex learning models.

To this end, in this paper we propose a hyperspectral face super-resolution (HFSR) method with limited training samples based on a spectral splitting and aggregation network (SSANet). Thanks to the amount of spectral bands of hyperspectral data, we split the hyperspectral image into different spectral groups and take each of them as an individual sample (in the sense that each group will be fed into the some network). By this splitting strategy, the training of the complex network can be guaranteed. To exploit the spectral correlations, which are destroyed by the splitting strategy but very important for the hyperspectral image reconstruction, we introduce an aggregation operator to gradually aggregate the neighbor bands at the deeper layers. As shown in Fig. \ref{fig:Network}, at the shallow layers, the complex network (with more blocks or channels) can be well-trained with `adequate' data (each group can be seen as a sample because of the parameter sharing strategy). At deeper layer, we design some light structures (with less blocks or channels) because we have reduced the training samples in order to exploit the spectral correlations. In general, although we only have a small number of training samples, we can expect to alleviate the S3 problem through this carefully designed spectral splitting and aggregation network.

Additionally, in order to further cope with the challenge of S3 problem, we propose to expand the training samples by self-representation learning. In particular, based on the assumption that there is a path from the training sample itself to the mean face of all training samples, we develop a training sample expanding strategy by self-representation learning. Given one training sample, we leverage the remaining training samples to represent it. According to a predefined smooth parameter, we can obtain a large number of synthetic samples, thus forming a path from the training sample itself to the mean face of the training set. In addition, we also expand the training dataset by the symmetry of the human face, thus we can directly double the size of the training dataset.

To sum up, this paper makes the following contributions: (i) We propose the first SR method for hyperspectral face image. (ii) A novel spectral splitting and aggregation network is introduced to ``generate'' more training samples to alleviate the S3 problem \emph{from less to more}, which is inevitable when we apply deep learning method to the HFSR task. (iii) We propose some schemes to expand the sample base based on self-representation and symmetry argumentation \emph{from less to more}.

\section{Related Work}
\label{sec:related}
Face SR is a very hot topic since the moment it was born. In recent years, a lot of methods have been proposed. In the following, we will revisit these approaches from global face based methods to local patch representation based methods, and then to the deep learning methods that have received much attention most recently. In addition, we also review some general hyperspectral image SR methods.

\subsection{Global Model-based FSR}
Face model based approaches leverage the global face statistical models, such as principal component analysis (PCA)~\cite{Wang2005Eig}, locality preserving projection (LPP) model~\cite{Park2007}, uniform space projection~\cite{Huang2010CCA, Anle2014face} and nonnegative matrix factorization (NMF)~\cite{Yang2010TIP}, to model the facial image and super-resolve the target HR face image globally. They can maintain the main structure of human face well. However, their reconstruction results lack detailed local facial features, and inevitably have ghosting artifacts.

\subsection{Local Patch-based FSR}
Considering that the human face is a highly structured subject, many face SR approaches try to exploit the prior knowledge by dividing the whole face image into small patches~\cite{Yang2013CVPR, li2014face, hui2017novel}. Among them, position-patch based methods have gained widespread attention in recent years. 
However, their solutions for the linear combination problem are unstable or not unique. To this end, our previous work~\cite{Jiang2014LcRTMM} introduces the locality-constrained representation (LcR) technique to simultaneously incorporate sparsity and locality into the patch representation. With the introduced locality constraint, it obtains a stable and reasonable representation. In order to alleviate the inconsistency of the two spaces, LR and HR spaces, some work have been proposed to iteratively obtain the patch representation and perform neighbor embedding or learn the mapping in correlation spaces~\cite{LiaoQM2015DSP, farrugia2015face, gao2020constructing, LiuLC2017Robust, shi2018hallucinating, chen2020robust}.


\subsection{Deep Learning-based FSR}
More recently, to reconstruct the latent HR image locally while thinking globally, DNNs, especially CNNs, have been applied to construct the mapping relationship between the LR images and their HR counterparts and shown strong learning capability and accurate prediction of HR images \cite{dong2016image, liu2016robust}. 
For example, Dong \emph{et al.}~\cite{dong2016image} developed a general image SR method based on SRCNN. This is the very first attempt to use deep learning tools for image SR reconstruction.
The approach of Liu \emph{et al.} \cite{liu2016robust} proposes to introduce the domain expertise to design a Sparse Coding based Network (SCN). Recently, R-DGN~\cite{yu2016ultra}, CBN~\cite{zhu2016deep}, LCGE~\cite{song2017learning}, Attention-FH \cite{Cao2017CVPR}, FSRNet \cite{CT-FSRNet-2018}, and \cite{Yu2018Imagining} are the most competitive approaches for face hallucination. They unitized very deep networks to model the relationship between the LR images and their HR counterparts, and verified that deeper networks can produce better results due to the large receptive field, which means considering more contextual information, \emph{i.e.}, very large image regions.

\subsection{Deep Learning-based HFSR}
According to whether an additional guided image (such as panchromatic, RGB, or multispectral image) is utilized, hyperspectral image SR techniques can be divided into two categories: multi-source information fusion based hyperspectral image SR (sometimes called hyperspectral image pansharpening) and single hyperspectral image SR \cite{loncan2015hyperspectral, yokoya2017hyperspectral}. The former is to leverage high-frequency spatial information from HR auxiliary image, and fuse them to the target HR hyperspectral image \cite{yokoya2011coupled, dong2016hyperspectral, xie2019multispectral, Borsoi2020TIP}. Though achieving good performance, they need a well co-registered auxiliary image which is arduous in real applications.
Without co-registered auxiliary image, the latter single hyperspectral image SR methods have still attracted considerable attention. How to exploit the abundant spectral correlations among successive bands is the essential problem for them. Traditional methods try to  incorporate the low-rank and group-sparse constraints to exploit the spectral-spatial prior \cite{huang2014super, li2016hyperspectral}. Due to its superior performance, deep learning techniques have also been introduced into the single
hyperspectral image SR task. They often adopt a two-step strategy to get an intermediate result through a deep network firstly, and then use spectral decomposition constraint to ensure the accuracy of the reconstructed spectral information \cite{yuan2017hyperspectral, xie2019hyperspectral}. Recently, there are also some approaches to directly learn an end-to-end deep networks to exploit the spatial-spectral prior. For example, Mei \emph{et al.} \cite{mei2017hyperspectral} presented a 3D full convolutional neural network to represent the spectral-spatial information. Wang \emph{et al}. \cite{li2020mixed} proposed to combine the advantages of 2D and 3D convolutions to exploit the spatial and spectral information. In \cite{li2018single}, Li \emph{et al.} developed a grouped deep recursive residual network (GDRRN) based single hyperspectral image SR method. The designed group-wise convolution and recursive structure can guarantee that it could yield very good performance. Inspired by the concept of group convolution, Jiang \emph{et al}. \cite{jiang2020learning} proposed the spatial-spectral prior network based super-resolution network (SSPSR). Most recently, inspired by the work of deep image prior \cite{ulyanov2018deep}, the approach of \cite{sidorov2019deep} presents an effective single hyperspectral image restoration algorithm. In general, these deep methods achieve better results than traditional methods. However, due to the limited hyperspectral training samples and the high dimensionality of spectral bands, it is difficult to fully exploit the spatial information and the correlation among the spectra of the hyperspectral data.

\textbf{\emph{Summary}}: As for the shallow learning based face SR methods, the local patch based models have much more strong representation ability than global face based models, rendering effectively the fine individual details to an input LR face. To obtain an accurate representation of the image patch, the current patch representation based methods all try to exploit the image priors (\emph{e.g}., collaborative, local, sparse, and low-rank constraints) or add some geometric regularizations from the HR space. They can well deal with the noisy input, inconsistency between LR and HR manifold spaces. These hand-designed prior models may not be effective. Deep neural network based methods leverage large-scale training dataset to train the network, and thus they can obtain very good performance. However, they may be not suitable for our task, where only about 100 samples are available. Although many SR methods for general hyperspectral images have been proposed in recent years, they all rely on a large-scale training set. Therefore, how to adapt the deep learning techniques for the HFSR task, especially when the training samples are very limited, is an urgent and extremely challenging issue.

\section{The Proposed Method}
\label{sec:proposedmethod}
Different from gray/RGB face SR problem, in which large paired HR and LR face training samples can be collected, HFSR can only leverage a small size of training dataset. Therefore, the very limited data cannot support the representation and modeling of complex hyperspectral data. To this end, in this paper we carefully design a deep network based on gradually spectral splitting and aggregation to alleviate the S3 problem. In addition, we also propose two strategies to expand the training samples by a self-representation model and the symmetry-induced augmentation. In this section, we will first present the details of SSANet and then introduce our training sample expanding strategies.


\subsection{Spectral Splitting and Aggregation Network}
\textbf{\emph{Spectral splitting module}}: As we discussed above, because the dimensionality of hyperspectral images is very high and the training samples are very limited, it is very difficult for existing deep neural network based approaches to effectively represent the hyperspectral data. Thanks to the amount of spectral bands of hyperspectral data, where each spectral band (neighbor spectral bands) can be seen as a training sample, we split the hyperspectral image into different groups (samples), thus largely expanding the size of the original training dataset. Note that we let each split group be fed into a branch network, and let different branches share the same parameters. Through this parameter sharing based splitting strategy, it will provide a strong data support for training a deep network.

As shown in Fig. \ref{fig:Network}, at the shallow layer (\emph{e.g.}, the first spectral splitting and aggregation stage) we split the hyperspectral data into as many groups as possible (a band is a group). In this way, we can obtain multiple training samples, which makes it possible to design a complex network (\emph{i.e.}, many spatial-spectral residual blocks (SSRBs)) to extract features of hyperspectral data. In the spectral splitting network, each group will be fed to a branch network to extract the features.

\textbf{\emph{Aggregation module}}: However, these extracted features may overlook the correlations among the spectral bands, which are very import to model and reconstruct the hyperspectral data. Therefore, at the deeper layers, we gradually increase the size (band number) of the group (\emph{i.e.}, the band number of each group, from 1 to 4, and then to 8, and finally to all 33 bands for different spectral splitting and aggregation stages). Meanwhile, we also let the neighbor groups overlap with each other. When we feed these overlapped groups to the splitting branch networks, we obtain the output by an aggregation operator, where the output is averaged according to their spectral indices. It should be noted that before feeding the outputs of different branch networks to the following aggregation module, we apply an additional \emph{Conv} layer to reduce the channel dimension to the number of input bands. Thus, each branch network acts like a ``reconstruction'' network and the input and output have the same channels. As we know, with the increase of the group size, the samples available for training will decrease. Thus, at the deeper layers we design some light structures (with less SSRBs). For example, with the progress of spectral splitting and aggregation reconstruction, the number of SSRBs decreases from 12 to 8, 4, and 2 for different spectral splitting and aggregation stages, respectively.

The proposed SSANet includes three spectral splitting and aggregation modules followed by a reconstruction module. In order to effectively and efficiently represent and reconstruct the hyperspectral face images, we carefully set the numbers of input band as well as the number of SSRB at each spectral splitting network. In addition, to fully exploit the spectral information of hyperspectral data, we also change the feature channels of different spectral splitting networks. At the shallow layer, where the input has only one spectral band, we set the number of feature channel to 64, and then increase it to 128, 128, and then to 192. This is because the input has more spectral bands with the progress of spectral splitting and aggregation reconstruction (from 1 to 4, 8, and then to 33), it calls for more feature channels to represent the spectral information of hyperspectral data.

\textbf{\emph{Upscaling strategy}}: For the proposed SSANet, another design worth mentioning is the upscaling strategy. The most commonly used upscaling strategy is the first upsampling (when the input is fed into the network) or the last upsampling (at the end of the network). In this paper, we are inspired by the pyramid SR network \cite{lai2017deep} and propose an upscaling search strategy. In particular, by inserting upscaling modules at different locations (before each aggregation module) of the network, we search for a set of optimized upscaling schemes for different upsampling factors. As shown in Fig. \ref{fig:Network}, above the input of each splitting network, we give the optimal upscaling scheme (and the size of the input image) under different upsampling factors, \emph{e.g.}, $\times 2$, $\times 4$, and $\times 8$. More details can be found at Section \ref{sec:upsampling}.

\textbf{\emph{Spatial-spectral residual block (SSRB)}}: Different from the commonly used residual module, we introduce a spatial-spectral residual block \cite{he2016deep}. It is a residual module like block, and includes one $3\times3$ \emph{Conv} and one $1\times1$ \emph{Conv} followed by an ReLU layer. We modify the residual module due to the following considerations. First, the added $1\times1$ \emph{Conv} can be leveraged to reorganize and reweight the importance of spectral bands, thus efficiently exploiting the spectral correlations of hyperspectral face image. Second, the added $1\times1$ \emph{Conv} can be seen as the purpose of information distillation and more useful information can be well preserved. Third, it deepens the depth of the network at a small cost in term of parameters.


\subsection{Training Sample Expanding}
As we discussed above, the spectral dimensionality of hyperspectral face image is very high, but the training sample number of available hyperspectral face image dataset is extremely small. Therefore, to well represent the high-dimensional hyperspectral face image, one possible solution is to expand the training set by synthesizing new training samples.

In this paper, we develop a self-representation learning method to synthesize the face image. In particular, we assume that all face images are from the same source, \emph{i.e.}, the mean face, and there is a path from the mean face to the individual face. 
Therefore, we can use some states on the path to obtain new samples. Mathematically, the self-representation based face synthesis method can be described as following,
\begin{equation}\label{eq:selfrepresentation}
\hat x_i = \sum_{j=1, j\not=i}^N w(x_i,x_j)x_j,
\end{equation}
where $x_i$ is the sample to be reconstructed, $\{x_j|j\not=i, 1\le j\le N\}$ are the training samples except $x_i$ itself, and $w(x_i,x_j)$ represents the reconstruction weight corresponding to $x_j$. From the above definitions, we can learn that the synthesized training sample is a linear combination of the training samples except itself. In particular, the combination weight is proportional to the similarity between current sample $x_i$ and the remaining training samples $x_j$,
\begin{equation}
w(x_i,x_j) = \frac{\text{sim}(x_i,x_j)}{\sum_{j\not=i,j=1}^N \text{sim}(x_i,x_j)}.
\end{equation}
Here, $\text{sim}(x_i,x_j)$ denotes the similarity between $x_i$ and $x_j$ and is defined as follow,
\begin{equation}\label{eq:similarity}
\text{sim}(x_i,x_j) = \exp\left(-\frac{||x_i-x_j||_2^2}{\sigma^2G} \right),
\end{equation}
where $\sigma$ denotes the synthesis controlling parameter, and $G$ is the mean of all elements of $\{||x_i-x_j||_2^2 |j\not=i, 1\le j\le N\}$. Since the entire face image is high-dimensional, the globally reconstruction methods may be not accurate or cannot introduce additional information to the training dataset (please refer to the \textbf{\emph{Remark 2}}). Thus, we decompose the global face into small image patches, and apply the above-mentioned approach for each small image patch. Lastly, by integrating all the reconstructed image patches, we can synthesize a new training sample.

\setlength{\tabcolsep}{0.50pt}
\begin{table}
\begin{center}
\caption{Effectiveness analysis of the proposed spectral splitting and aggregation structure. }\label{tab:SSAS}
\footnotesize
\begin{tabular}{c|c|cccccc}
\hline
&Dataset  &   CC$\uparrow$	&	SAM$\downarrow$	&	RMSE$\downarrow$	&	ERGAS$\downarrow$	&	PSNR$\uparrow$	&	SSIM$\uparrow$\\
\hline
\hline
\multirow{4}{*}{w/o SSAS} & O	&	.9987 	&	.8100 	&	.0054 	&	.4780 	&	46.6650 	&	.9911 	\\
&O+Self	&	.9990 	&	.7088 	&	.0048 	&	.4179 	&	47.9166 	&	.9928 	\\
&O+Sym	&	.9990 	&	.7271 	&	.0049 	&	.4279 	&	47.6540 	&	.9925 	\\
&O+Self+Sym	&	.9991 	&	.6834 	&	.0046 	&	.4041 	&	48.2217 	&	.9932 	\\
\hline
\multirow{4}{*}{SSAS} & O	&	.9989 	&	.7555 	&	.0051 	&	.4440 	&	47.2672 	&	.9923 	\\
&O+Self	&	.9992 	&	.6552 	&	.0044 	&	.3825 	&	48.6323 	&	.9938 	\\
&O+Sym	&	.9992 	&	.6549 	&	.0043 	&	.3746 	&	48.7505 	&	.9941 	\\
&O+Self+Sym	&	.9993 	&	.6086 	&	.0039 	&	.3430 	&	49.4840 	&	.9948 	\\
\hline
\multicolumn{2}{c|}{Average Improvements}           &   .0002$\uparrow$ 	&	.0638$\downarrow$ 	&	.0005$\downarrow$ 	&	.0460$\downarrow$ 	&	.9192$\uparrow$ 	&	.0014$\uparrow$\\
\hline
\end{tabular}
\normalsize
\end{center}
\end{table}

\textbf{\emph{Remark 1}}: The human face, as a highly structured object, has two eyes, a nose, and a mouth. The global structure information of different objects is very similar. The differences between different people are mainly reflected in the detailed features. Therefore, we have reason to believe that through self-representation learning, we can synthesize some feature detail information that is not available in the training dataset. This information can actually be regarded as a kind of noise compared to the represented face image. Therefore, the proposed method based on self-representation learning can bring more information to the training dataset.

\textbf{\emph{Remark 2}}: It seems that putting a reconstructed sample back to the training set will add no extra information to the dataset. If we reconstruct the face image globally, the above point is the truth. But when we reconstruct the face image locally (patch-wisely), this would be not true. In other words, for those patch-based methods, the reconstructed HR result does not have to be a linear combination of all the original training samples. We can imagine an extreme situation: if we set the patch size to very small level, \emph{i.e}., pixel-wise, we can reconstruct any image (any content) and dose not have to be a linear combination of training samples, \emph{e.g}., a cat or a dog image. Therefore, these patch-based reconstruction method can reconstruct some samples, which dose not have to be a linear combination of the original training samples. Therefore, when we put reconstructed HR samples back to the training set, we can actually introduce some additional information in the sense of that we can add an image (which cannot be linear combination with the original training) to the training dataset.

\section{Experiments}
\label{sec:experiments}
We use Pytorch libraries\footnote{\url{https://pytorch.org}} to implement and train the proposed SSANet method (the code will be available upon acceptance). We train different models to super-resolve the hyperspectral face images for upsampling factors 2, 4 and 8 with random initialization. We use the ADAM optimizer \cite{kingma2014adam} with an initial learning rate of 1e-4 which decays by a factor of 10 at each 30 epochs. In our experiments, we find it will take 60 epochs to achieve the stable performance. The models are trained with a batch size of 4.

\begin{figure*}
  \centering
  \includegraphics[width=0.98\linewidth]{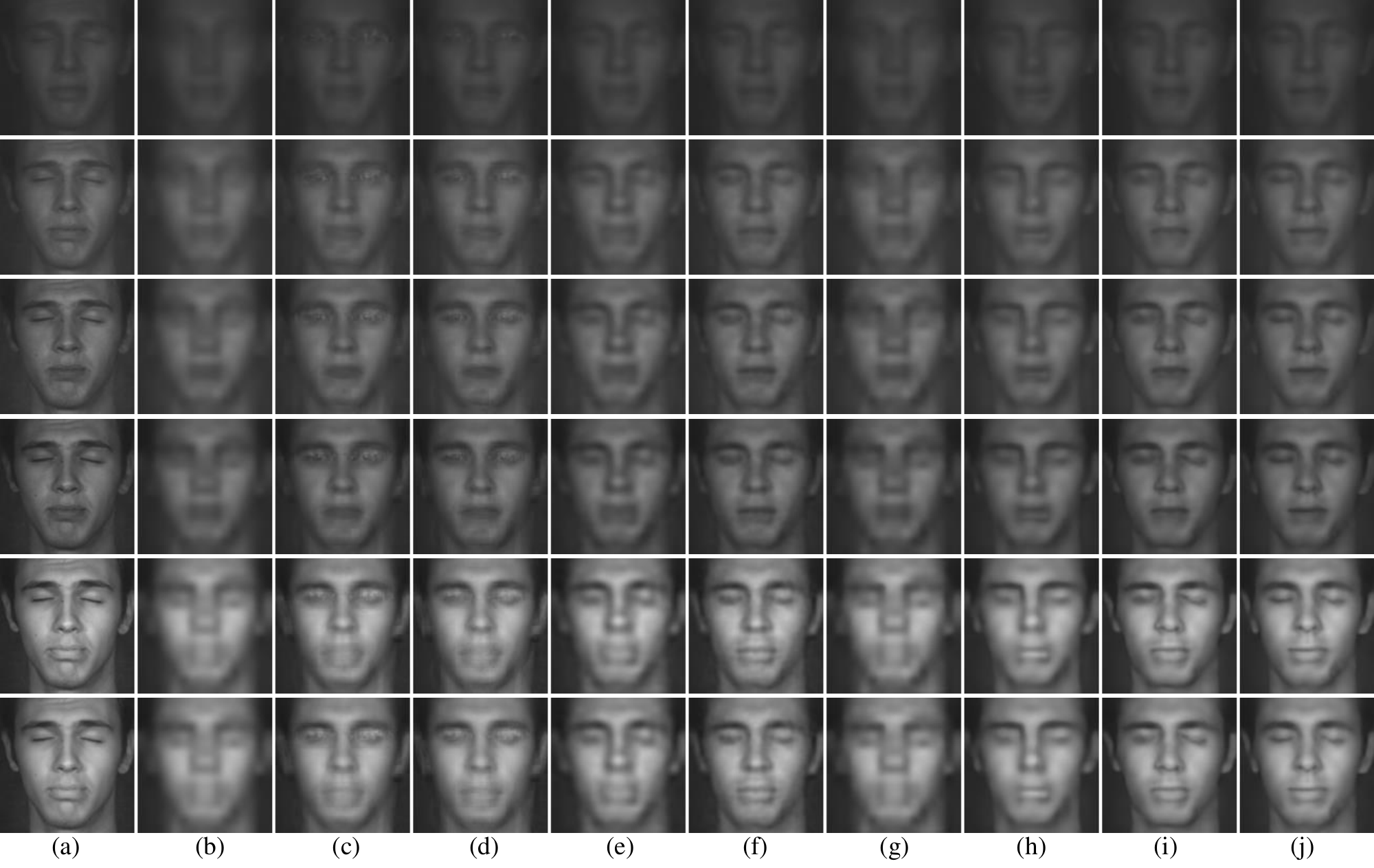}\\
  \vspace{-0.35cm}
  \caption{8 $\times$ SR results of one object of different methods at bands 5, 10, 15, 20, 25, and 30. (a) HR faces, (b) Bicubic interpolation, (c) results of LSR \cite{Ma2010LSR}, (d) results of LcR \cite{Jiang2014LcRTMM}, (e) results of EDSR \cite{lim2017enhanced}, (f) results of SAN \cite{dai2019second}, (g) results of 3DCNN \cite{mei2017hyperspectral}, (h) results of GDRRN \cite{li2018single}, (i) results of SSPSR \cite{jiang2020learning}, (j) Our results.}\label{fig:SingleResults}
\end{figure*}

All the experiments are conducted on the UWA Hyperspectral Face Database (UWA-HSFD)\footnote{It should be noted that we only use the database of UWA-HSFD due to the following reason. Currently, there are only three widely used hyperspectral face databases, the Hong Kong Polytechnic University Hyperspectral Face Database (PolyU-HSFD) \cite{di2010PolyU}, the CMU Hyperspectral Face Database (CMU-HSFD) \cite{Denes2002}, and UWA-HSFD used in our paper. The face images in PolyU-HSFD are very noisy and not suitable for the supervised learning. As for CMU-HSFD, the related project is unavailable at present, and we cannot obtain the database.} \cite{uzair2015hyperspectral}. 
It contains 145 hyperspectral face images of 80 subjects collected in four sessions over time. The face images are acquired by an indoor imaging system using a CRI's VariSpec LCTF filter integrated with a Photon focus camera. In UWA-HSFD \cite{uzair2015hyperspectral}, the spectral range, in which the images have been acquired, is from 400 nm to 720 nm with a step size of 10 nm, resulting 33 spectral bands. Alignment errors are present between individual bands due to subjects movements and eye blinking during image acquisition.
To simulate the situation that the testing face image is not in the training phases, we randomly selecting one cube from one session for each of the 70 subjects (includes 10\% evaluation samples). The remaining 10 subjects are used for testing. In our experiments, all the input LR images are obtained by 2$\times$, 4$\times$, or 8$\times$ Bicubic downsampling.

\textbf{Evaluation measures}. Six widely used quantitative picture quality indices (PQIs) are employed to evaluate the performance of our method, including cross correlation (CC) \cite{loncan2015hyperspectral}, spectral angle mapper (SAM) \cite{yuhas1992discrimination}, root mean squared error (RMSE), erreur relative globale adimensionnelle de synthese (ERGAS) \cite{wald2002data}, peak signal-to-noise ratio (PSNR), and structure similarity (SSIM) \cite{wang2004image}. For PSNR and SSIM of the reconstructed hyperspectral images, we report their mean values of all spectral bands. CC, SAM, and ERGAS are three widely adopted quality indices in HS fusion task, while the remaining three indices are commonly used quantitative image restoration quality indices. The best values for these indices are 1, 0, 0, 0, $+\propto$, and 1, respectively.


\subsection{Investigation to the Proposed SSANet}
\subsubsection{Progressively upsampling}
\label{sec:upsampling}
The simplest upsampler is to do upsampling at the beginning by Bicubic interpolation or perform pixel shuffle at the end. They either increase the parameters and computational complexity of the network or increase the difficulty of neural network training.
Inspired by the Laplacian pyramid SR network \cite{lai2017deep}, we perform upsampling at the intermediate layers to progressively upsample the input image. We insert the upsampling module to each spectral splitting and aggregation module, as shown in Fig. \ref{fig:Network}. Similar to the Neural Architecture Search (NAS) \cite{liu2018progressive}, in our experiments we manually search the optimal structure. More details can be found \emph{in the supplementary}.


\setlength{\tabcolsep}{2.5pt}
\begin{table}
\begin{center}
\caption{Effectiveness analysis of the proposed training sample expanding strategy. }\label{tab:trainingsample}
\footnotesize
\begin{tabular}{cccccccc}

  \hline
Dataset & $r$  &   CC$\uparrow$	&	SAM$\downarrow$	&	RMSE$\downarrow$	&	ERGAS$\downarrow$	&	PSNR$\uparrow$	&	SSIM$\uparrow$ \\
\hline
O	&	2	&	0.9989 	&	0.7555 	&	0.0051 	&	0.4440 	&	47.2672 	&	0.9923 	\\
O+Self	&	2	&	0.9992 	&	0.6552 	&	0.0044 	&	0.3825 	&	48.6323 	&	0.9938 	\\
O+Sym	&	2	&	0.9992 	&	0.6549 	&	0.0043 	&	0.3746 	&	48.7505 	&	0.9941 	\\
O+Self+Sym	&	2	&	0.9993 	&	0.6086 	&	0.0039 	&	0.3430 	&	49.4840 	&	0.9948 	\\
\hline
O	&	4	&	0.9957 	&	1.2985 	&	0.0110 	&	0.9269 	&	41.4541 	&	0.9673 	\\
O+Self	&	4	&	0.9960 	&	1.2496 	&	0.0107 	&	0.9002 	&	41.8074 	&	0.9687 	\\
O+Sym	&	4	&	0.9958 	&	1.2734 	&	0.0109 	&	0.9195 	&	41.6118 	&	0.9678 	\\
O+Self+Sym	&	4	&	0.9960 	&	1.2409 	&	0.0107 	&	0.9003 	&	41.8483 	&	0.9689 	\\
\hline
O	&	8	&	0.9825 	&	2.0828 	&	0.0230 	&	1.8961 	&	35.2369 	&	0.8901 	\\
O+Self	&	8	&	0.9838 	&	1.9939 	&	0.0221 	&	1.8281 	&	35.6496 	&	0.8956 	\\
O+Sym	&	8	&	0.9828 	&	2.0371 	&	0.0228 	&	1.8804 	&	35.3614 	&	0.8909 	\\
O+Self+Sym	&	8	&	0.9842 	&	1.9934 	&	0.0219 	&	1.8095 	&	35.6852 	&	0.8973 	\\
\hline
\end{tabular}
\normalsize
\end{center}
\end{table}

\begin{figure*}[!t]
  \centering
  \includegraphics[width=0.98\linewidth]{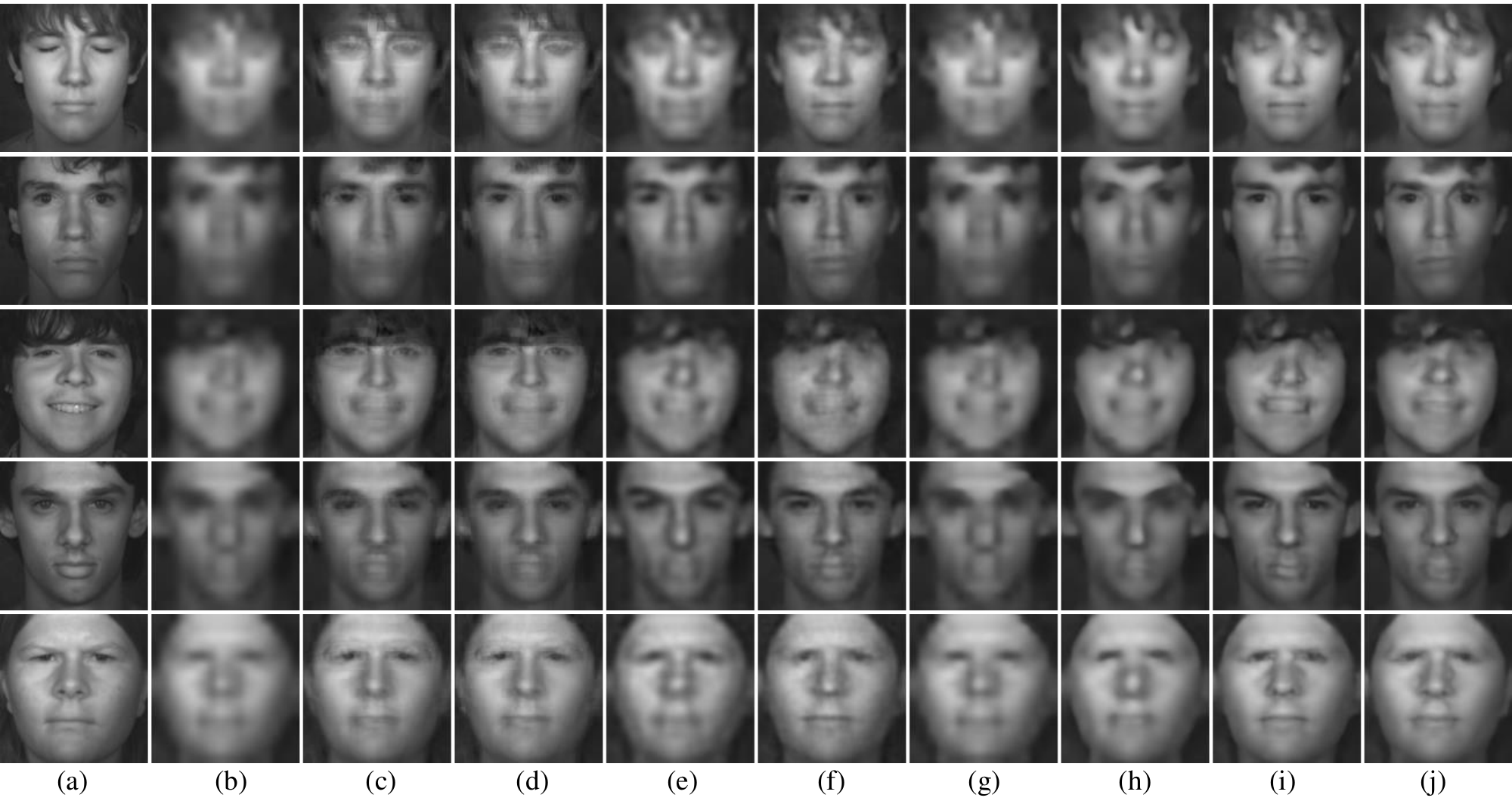}\\
  \vspace{-0.35cm}
  \caption{8 $\times$ SR results of five objects of different methods at band 30. The meanings of (a)-(j) are the same as Fig. \ref{fig:SingleResults}.}\label{fig:AllResults}
\end{figure*}

\subsubsection{Spectral splitting and aggregation}
To alleviate the S3 problem, we carefully design the spectral splitting and aggregation strategy (SSAS). It splits the hyperspectral face image into as many groups as possible at shallow layers to guarantee the training of a complex network, and gradually aggregates the neighbor bands at deeper layers to fully exploit spectral correlations. To demonstrate the effectiveness of SSAS, we report the comparison results of the proposed deep neural network with and without SSAS in Table \ref{tab:SSAS} when the upsampling factor is 2. To simulate the network without SSAS, we maintain the primary structure of SSANet and set the band size of each branch network to 33 (the number of spectral bands for UWA-HSFD database). In this situation, there is no spectral splitting and aggregation process for SSANet. From the tabulated results, we can clearly learn that SSANet with SSAS can stably improve the performance regardless of the training datasets (the original training dataset and expanded training dataset). In all cases, the average improvement (in term of PSNR) of SSANet with SSAS over SSANet without SSAS is nearly 1.0 dB, and the spectral similarity has also increased by nearly 10\% (please refer to the SAM index). 


\subsection{Effectiveness of Training Sample Expanding}
In this paper, we introduce two strategies to expand the size of training dataset. To demonstrate the effectiveness of these strategies, we show the comparison results of our method with different training datasets. As shown in Table \ref{tab:trainingsample}, ``O'' denotes the original training dataset, ``Self'' demotes the expanded training dataset with self-representation based data argumentation, and ``Sym'' is the expanded training dataset using the symmetry nature of humane to flip the face image to double the size of training dataset. From the comparison results, we can clearly learn that the proposed self-representation and symmetry based data argumentation methods can improve the performance of our proposed deep neural network separately, no matter what the upsampling factor $r$ is. The combination of these two strategies can achieve the best performance.

\subsection{Comparison Results}
Since the proposed SSANet is the first hyperspectral image SR method for human faces, there is no direct hyperspectral image SR method for comparison. Therefore, to demonstrate the effectiveness of our method, we adjust two traditional shallow learning based face SR methods, LSR \cite{Ma2010LSR} and LcR \cite{Jiang2014LcRTMM}, two deep learning based general image SR methods, EDSR \cite{lim2017enhanced} and SAN \cite{dai2019second}, and one most recently proposed deep learning based single hyperspectral image SR method, SSPSR \cite{jiang2020learning} for the hyperspectral image SR task. Specifically, as for LSR \cite{Ma2010LSR} and LcR \cite{Jiang2014LcRTMM}, we super-resolve each band of the hyperspectral face image separately. For EDSR \cite{lim2017enhanced} and SAN \cite{dai2019second}, we retrain their networks by changing the number of input channels with the UWA-HSFD database. SSPSR \cite{jiang2020learning} is also retrained by the UWA-HSFD database for fair comparison.

Fig. \ref{fig:SingleResults} shows the super-resolved results of one object of different SR approaches at five specific bands, and in Fig. \ref{fig:AllResults} we also present the results of five objects at band 30. From these comparison results, we learn that Bicubic interpolation lost most of the face detailed features. LSR \cite{Ma2010LSR} and LcR \cite{Jiang2014LcRTMM} require the accurate alignment of face images. However, face images cannot be completely aligned due to non-rigid transformations such as expressions and poses. Therefore, ghosting effects often appear at the regions where the eyes and mouth cannot be completely aligned. EDSR \cite{lim2017enhanced} and SAN \cite{dai2019second} are not specifically designed for hyperspectral faces, and they cannot fully exploit the rich spectral information of hyperspectral faces. Their results are much blur than SSPSR \cite{jiang2020learning}, which is a general hyperspectral image SR method. Results reconstructed by our proposed method are more credible (more similar to the ground truth), and at the same time our method greatly reduces the artificial effects of the reconstructed eyes, mouth, and nose.

It should be noted that almost all these comparison approaches cannot accurately reconstruct the slightly opened eyes. Actually, this shows a weakness of these learning-based techniques which depend highly upon the consistency (similarity) between the supporting testing image and training dataset. This is mainly because the training data cannot cover the changes caused by all factors (such as posture, expression, lighting, \emph{etc}.), especially when the size of the training dataset is limited. 

Table \ref{tab:Face} tabulates the average objective results of seven comparison methods on the 10 testing images from UWA-HSFD dataset when the upsampling factor is set to 2, 4, and 8. The best and second best results are highlighted in boldface and underlined, respectively. Note that SSANet is the model trained by the original training samples, while SSANet+ denotes the model trained by the expanded training samples with ``Self'' and ``Sym''. These objective results also demonstrate the better performance of our method.

\setlength{\tabcolsep}{2.5pt}
\begin{table}
\begin{center}
\caption{Quantitative comparisons of different approaches.}\label{tab:Face}
\footnotesize
\begin{tabular}{cccccccc}

\hline
	&	$r$	&	CC$\uparrow$	&	SAM$\downarrow$	&	RMSE$\downarrow$	&	ERGAS$\downarrow$	&	PSNR$\uparrow$	&	SSIM$\uparrow$	\\
\hline
\hline
LSR \cite{Ma2010LSR}	&	2	&	0.9977 	&	1.2394 	&	0.0076 	&	0.6526 	&	43.7839 	&	0.9784 	\\
LcR	\cite{Jiang2014LcRTMM} &	2	&	0.9980 	&	1.0751 	&	0.0073 	&	0.6214 	&	44.4102 	&	0.9808 	\\
EDSR \cite{lim2017enhanced}	&	2	&	\underline{0.9988} 	&	\underline{0.7777} 	&	0.0053 	&	0.4616 	&	\underline{46.9760} 	&	\underline{0.9917} 	\\
SAN	\cite{dai2019second} &	2	&	0.9985 	&	0.8484 	&	0.0055 	&	0.4921 	&	46.3874 	&	0.9906 	\\
3DCNN \cite{mei2017hyperspectral}	&	2	&	0.9986	&	0.8676	&	0.0061	&	0.5218	&	46.2634	&	0.9895	\\
GDRRN \cite{li2018single}	&	2	&	0.9976	&	0.7848	&	0.0054	&	0.4730	&	46.8245	&	0.9911	\\
SSPSR \cite{jiang2020learning}	&	2	&	0.9987 	&	0.7947 	&	\underline{0.0052} 	&	\underline{0.4614} 	&	46.9406 	&	0.9916 	\\
SSANet	&	2	&	\textbf{0.9989} 	&	\textbf{0.7555} 	&	\textbf{0.0051} 	&	\textbf{0.4440} 	&	\textbf{47.2672} 	&	\textbf{0.9923} 	\\
\hline
LSR \cite{Ma2010LSR}	&	4	&	0.9937 	&	1.6073 	&	0.0134 	&	1.1214 	&	39.6160 	&	0.9472 	\\
LcR	\cite{Jiang2014LcRTMM} &	4	&	0.9939 	&	1.5774 	&	0.0134 	&	1.1144 	&	39.7621 	&	0.9480 	\\
EDSR \cite{lim2017enhanced}	&	4	&	0.9949 	&	1.4242 	&	0.0118 	&	0.9995 	&	40.6932 	&	0.9628 	\\
SAN	\cite{dai2019second} &	4	&	0.9948 	&	1.4723 	&	0.0118 	&	1.0048 	&	40.6052 	&	0.9629 	\\
3DCNN \cite{mei2017hyperspectral}	&	4	&	0.9942	&	1.4482	&	0.0132	&	1.0984	&	40.0966	&	0.9556	\\
GDRRN \cite{li2018single}	&	4	&	0.9950	&	1.3897	&	0.0118	&	0.9939	&	40.7684	&	0.9626	\\
SSPSR \cite{jiang2020learning}	&	4	&	\underline{0.9953} 	&	\underline{1.3750} 	&	\underline{0.0114} 	&	\underline{0.9656} 	&	\underline{41.0340} 	&	\underline{0.9651} 	\\
SSANet	&	4	&	\textbf{0.9957} 	&	\textbf{1.2985} 	&	\textbf{0.0110} 	&	\textbf{0.9269} 	&	\textbf{41.4541} 	&	\textbf{0.9673} 	\\
\hline
LSR	\cite{Ma2010LSR} &	8	&	0.9795 	&	2.3873 	&	0.0246 	&	2.0349 	&	34.4513 	&	0.8782 	\\
LcR	\cite{Jiang2014LcRTMM} &	8	&	0.9810 	&	2.3680 	&	0.0240 	&	1.9775 	&	34.8670 	&	0.8835 	\\
EDSR \cite{lim2017enhanced}	&	8	&	\underline{0.9820} 	&	2.2191 	&	\underline{0.0232} 	&	\underline{1.9213} 	&	\underline{35.1152} 	&	0.8849 	\\
SAN	\cite{dai2019second} &	8	&	0.9815 	&	2.3889 	&	0.0234 	&	1.9420 	&	34.9323 	&	0.8855 	\\
3DCNN \cite{mei2017hyperspectral}	&	8	&	0.9741	&	2.2846	&	0.0283	&	2.3150	&	33.5437	&	0.8470	\\
GDRRN \cite{li2018single}	&	8	&	0.9785	&	\underline{2.1964}	&	0.0254	&	2.0910	&	34.3277	&	0.8704	\\
SSPSR \cite{jiang2020learning}	&	8	&	0.9816 	&	2.1979 	&	0.0234 	&	1.9356 	&	34.9966 	&	\underline{0.8856} 	\\
SSANet	&	8	&	\textbf{0.9825} 	&	\textbf{2.0828} 	&	\textbf{0.0230} 	&	\textbf{1.8961} 	&	\textbf{35.2369} 	&	\textbf{0.8901} 	\\
  \hline
\end{tabular}
\normalsize
\end{center}
\end{table}

\section{Conclusions}
\label{sec:refs}
In this paper, we present a hyperspectral face super-resolution (HFSR) approach based on spectral splitting and aggregation network (SSANet). This is the first face SR work focusing on hyperspectral images. Different from traditional face and general image SR tasks, in which there are enough training samples to support the training of a complex network, HFSR has to face the small training sample size (S3) problem. To this end, on the one hand we carefully design a spectral splitting and aggregation network to ``generate'' more training samples to alleviate the S3 problem \emph{from less to more}, and simultaneously to fully make use of multiple spectral information. On the other hand we introduce two strategies to expand the training samples by a self-representation and the symmetry augmentation. Experimental results on public hyperspectral face database demonstrate that our proposed SSANet method and the self-representation and symmetry based training sample expanding strategy are effective for the HFSR task. In addition, we also report the comparison results on hyperspectral images of natural scenes, which demonstrate the generalization of our method.

\noindent\textbf{Acknowledgments:} The research was supported by the National Natural Science Foundation of China (61971165), in part by the Fundamental Research Funds for the Central Universities (FRFCU5710050119), the Natural Science Foundation of Heilongjiang Province (YQ2020F004).

{\small
\bibliographystyle{ieee_fullname}
\bibliography{egbib}
}

\end{document}